\documentclass{article}

\pdfoutput=1
    \PassOptionsToPackage{numbers, compress}{natbib}

     \usepackage[final]{neurips_2020}

\usepackage[utf8]{inputenc} 
\usepackage[T1]{fontenc}    
\usepackage{hyperref}       
\usepackage{url}            
\usepackage{booktabs}       
\usepackage{amsfonts}       
\usepackage{nicefrac}       
\usepackage{microtype}      
\usepackage{graphicx}
\usepackage{multirow}
\usepackage[all]{hypcap}

\title{Investigating Learning in Deep Neural Networks using Layer-Wise Weight Change}

\author{%
  Ayush Manish Agrawal\thanks{Equal Contributors}\hspace{0.2cm}\footnotemark[2]\\
  University of Nebraska\\
  \texttt{aagrawal@nebraska.edu} \And
  Atharva Tendle\footnotemark[1]\hspace{0.2cm}\thanks{OpenMined $\&$ Manifold Computing}\\
  University of Nebraska-Lincoln\\
  \texttt{atharva.tendle@huskers.unl.edu}\And
  Harshvardhan Sikka\footnotemark[2]\\
  Georgia Institute of Technology\\
  \texttt{harsh@manifold.com}\And
  Sahib Singh\footnotemark[2]\\
  Ford R\&A\\
  \texttt{sahibsingh570@gmail.com}\And
  Amr Kayid\footnotemark[2]\\
  German University in Cairo\\
  \texttt{amrmkayid@gmail.com}
   \\
}

\begin{document}

\maketitle

\begin{abstract}
Understanding the per-layer learning dynamics of deep neural networks is of significant interest as it may provide insights into how neural networks learn and the potential for better training regimens. We investigate learning in Deep Convolutional Neural Networks (CNNs) by measuring the relative weight change of layers while training. Several interesting trends emerge in a variety of CNN architectures across various computer vision classification tasks, including the overall increase in relative weight change of later layers as compared to earlier ones.
\footnote{ Code available at: \href{https://github.com/Manifold-Computing/Layer-Wise-Learning-Trends-PyTorch}{https://github.com/Manifold-Computing/Layer-Wise-Learning-Trends-PyTorch}}
\end{abstract}

\section{Introduction}
Deep learning based approaches have achieved excellent performance in a variety of problem areas, and generally consist of neural network based models that learn mappings between task specific data and corresponding solutions. The success of these methods relies on their ability to learn multiple representations at different levels of abstraction, achieved through the composition of non-linear modules that transform incoming representations into new ones \cite{lecun2015deep}. These transformation modules are referred to as layers of the neural network, and neural networks with several such layers are referred to as deep neural networks. Significant research has demonstrated the capacity for deep networks to learn increasingly complex functions, often through the use of the specific neural network primitives that introduce information processing biases in the problem domain. For example, in the vision domain, Convolutional Neural Networks (CNNs) utilize convolution operations that use filtering to detect local conjunctions of features in images, which often have local values that are highly correlated and invariant to location in the image. 

Various approaches have emerged that take advantage of the learning behavior of deep neural networks to improve their computational cost or reliability through interpretation. For example, transfer learning is a paradigm that focuses on transferring knowledge across domains, and often involves fine tuning neural networks that have been previously trained in a related domain to solve a new target task. This offers several advantages over training new networks from scratch on the task, as the prior learned parameters allow the network to learn the new task faster, assuming the pretraining domain is similar to the new one. Alongside this, a general observation in many computer vision tasks is that early layers converge to simple feature configurations \cite{yosinski2014transferable}. This phenomena is observed in many vision architectures, including Inception and Residual Networks \cite{cammarata2020thread:}. These findings, among others, point to a natural question: \textit{Do different layers in neural networks converge to their learned features at different times in the training process?} 

Understanding the layer-wise learning dynamics that allow for a deep neural network to learn the solution of a particular task is of significant interest, as it may provide insight into understanding potential areas of improvement for these algorithms and reduce their overall training costs. In this work, we empirically investigate the learning dynamics of different layers in various deep convolutional neural network architectures on several different vision tasks.

Our contributions are as follows:

\begin{itemize} 
    \item A metric to track the relative weight change in a given neural network layer on an epoch by epoch basis. We present relative weight change as a proxy for layer-wise learning, with the assumption that when the weights of a network have minimal change over a set of epochs, they are converging to their optimum.
    \item We track the relative weight change of several popular convolutional neural network architectures, including ResNets, VGG, and AlexNet for four benchmark datasets, including CIFAR-10, CIFAR-100, MNIST, and FMNIST. 
    \item Learning dynamics are analyzed from the perspective of relative weight change for complex and simple learning tasks for shallow and deep networks with different architectural motifs. Several key trends emerge, including early layers exhibiting less relative weight change than later layers over the course of training across the CNN architectures.
\end{itemize}

The rest of this text is organized as follows: Section 2 presents related work. Section 3 introduces relative weight change and our experimental methodology. Section 4 discusses empirical results across several datasets and architectures. Finally, Section 5 discusses conclusions and future directions for this line of research.

\section{Related Work}
While Deep Learning explainability is an active area of research, there has been limited research understanding the layer level trends in neural networks. Most of the work done so far in this context has been focused towards Feature Visualization of Neural Networks \cite{li2020visualizing, Erhan2009VisualizingHF, simonyan2014deep, nguyen2016multifaceted, nguyen2019understanding, zeiler2014visualizing, olah2017feature, Szegedy_2015_CVPR}. While our work is similar in context, our work presents a novel approach of understanding these features-  Instead of focusing on feature visualization we focus on understanding the weights in each layer of the neural network and compute the relative change in these weights across epochs. We then investigate these trends to various architectures (AlexNet, VGG-19, and ResNet-18 Network) and analyze the commonalities of these trends.

\section{Methods}

\subsection{Relative Weight Change}
To better understand the layer-wise learning dynamics through the training process, we introduce a metric known as Relative Weight Change (RWC). RWC can be understood to represent the average of the absolute value of the percent change in the magnitude of a given layer's weight. It can be formalized as

\begin{equation} 
\label{RWC}
\centering 
RWC_{L} = \frac{||w_{t} - w_{t-1} ||_{1}}{|| w_{t-1}||_{1}}
\end{equation}
where $L$ represents a single layer in a deep neural network, and $w_{t}$ represents the vector of weights associated with $L$ at a given training step $t$. We use the $L_{1}$ norm to characterize the difference in magnitude of the weights, and normalize the difference by dividing by the magnitude of the layer's weights during the previous training step. Following this, an averaging step is applied to get a single value for RWC across the entire layer. The resulting proportion informs us as to how much the layer's weights are changing over training steps. Smaller changes over a prolonged period indicate that the layer's weights are nearing an optimum. We use this measure to characterize weight dynamics as on a per-layer basis as a function of training iterations to better understand how layers are learning.

\subsection{Experimental Approach}

\textbf{Datasets and Settings} We use four benchmark datasets: CIFAR-10 \cite{Krizhevsky09learningmultiple} which contains 60,000 images of 10 classes, CIFAR-100 \cite{Krizhevsky09learningmultiple} which contains 60,000 images of 100 classes, MNIST Handwritten Digits \cite{lecun-mnisthandwrittendigit-2010}, and FMNIST Fashion-MNIST \cite{xiao2017/online} that contains 60,000 images of 10 classes. These benchmark datasets see significant use in deep learning research. The datasets also provide good variety in the complexity of their associated learning tasks. MNIST is fairly easy for simple networks to solve, FMNIST and CIFAR-10 provide new levels of complexity in image content and detail, and CIFAR-100 has significantly more classes and fewer samples per class, ramping up difficulty considerably. 

\textbf{Network Structure and Training} 
We use ResNet18 \cite{he2016deep}, VGG-19 with Batch Norm \cite{simonyan2014very}, and AlexNet \cite{krizhevsky2017imagenet}. These architectures were chosen for several reasons. They are ubiquitously used in the research community for computer vision problems. They also provide some variety in the information processing techniques and biases utilized to learn from images. For example, ResNets make use of residual connections, skip connections, and blockwise design while VGG makes use of significant downsampling and depth. A variety of architectures is useful for establishing some of the general trends we oberve in this work, and inconsistencies may be attributable to the concrete differences between them. They also represent a good distribution of computational complexity, as AlexNet is significantly shallower than both ResNet and VGG variants. 

The general training strategies used for these architectures was mostly consistent with those demonstrated in their respective papers. It's worth noting that the state of the art accuracy on our datasets required adaptive learning rate. We made a decision to exclude that from our training to focus on the layer-wise learning patterns in these deep networks. We used Stochastic Gradient Descent (SGD) \cite{Kiefer1952StochasticEO} with momentum and weight decay for our experiments. The learning rate was kept constant throughout the experiments and each model was trained for a total of 150 epochs. Table 1, included in the supplementary material, shows the detailed hyperparameters used for training on different architectures.

To interpret the layer-wise learning, we find the RWC as formulated in \ref{RWC} for each layer per epoch. We run the same experiment for each architecture with different weight initializations using 5 different seeds to reduce the possibility of observing trends specific to a single run. We store the RWC array from each experiment, plot the average of the associated curves, and report the results in the following section.

\section{Results}
Here, we include empirical results and analyses of layer-wise weight changes collected through the experimental approach described previously. Results are broken down by overall architecture, with trends highlighted for each of the 4 datasets. Figures demonstrating the RWC of specific layers are included and referenced in each set of analyses. 

\subsection{Residual Networks}
The ResNet architecture is a deep convolutional network that consists of a repeated block motif of convolutional and batch normalization layers, along with residual connections between early and later layers. The convolutional hyperparameters of blocks are standardized. ResNet-18, used in these experiments, consists of 4 such residual blocks which we track explicitly as part of our analyses. 

\begin{figure}[hbt!]
    \centering
    \noindent\makebox[\textwidth]{\includegraphics[width=15cm]{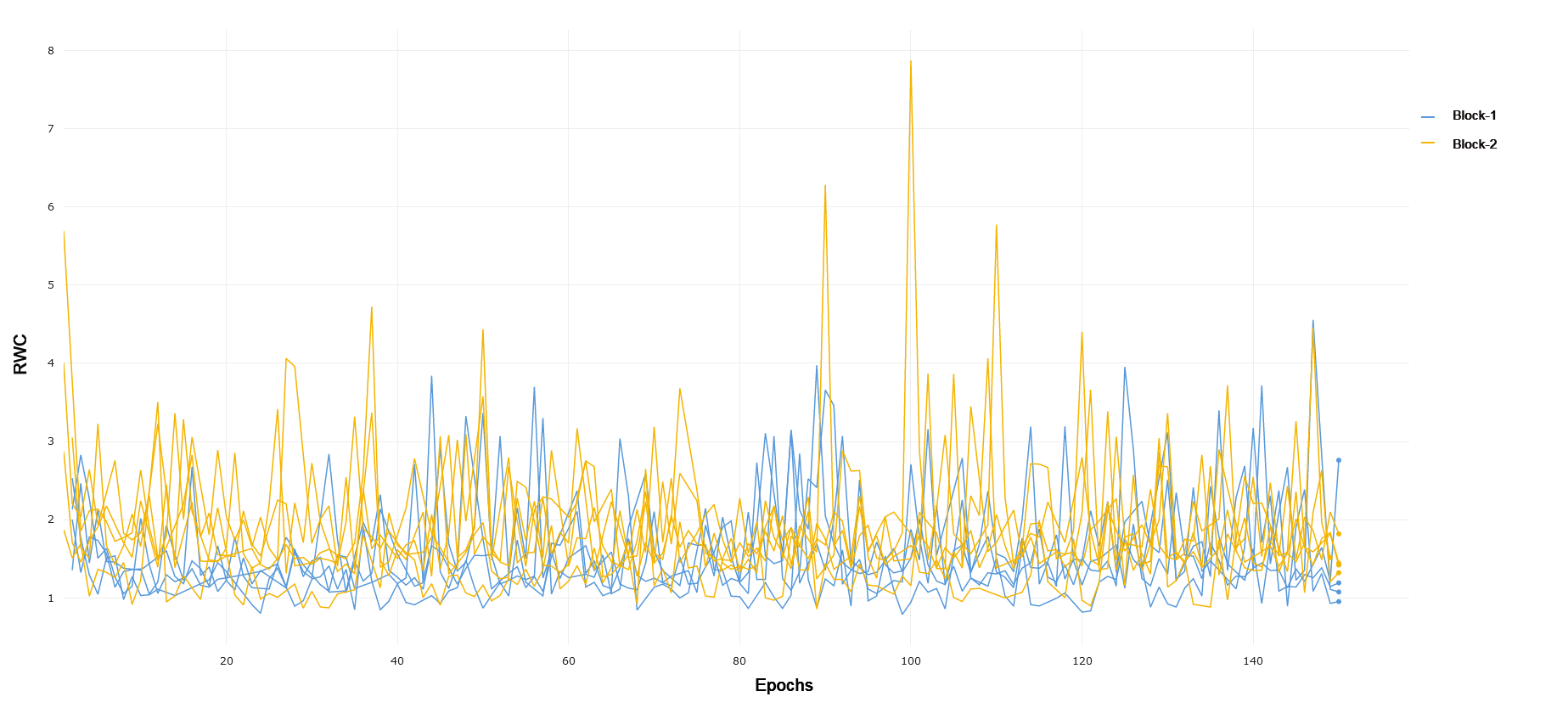}}
    \caption{RWC for ResNet-18 Blocks 1-2 on CIFAR-10}
    \label{fig:0}
\end{figure}

\begin{figure}[hbt!]
    \centering
    \noindent\makebox[\textwidth]{\includegraphics[width=15cm]{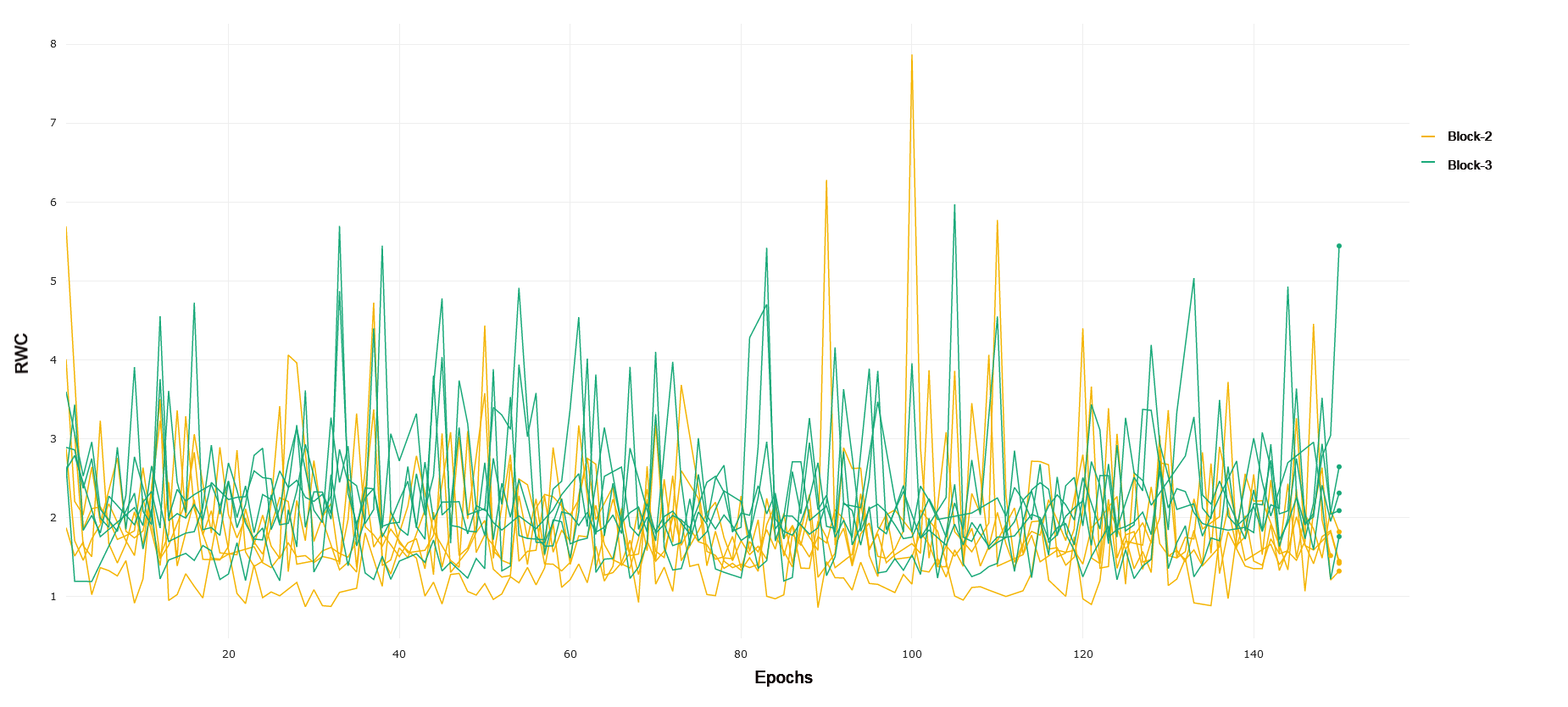}}
    \caption{RWC for ResNet-18 Blocks 2-3 on CIFAR-10}
    \label{fig:1}
\end{figure}

\begin{figure}[hbt!]
    \centering
    \noindent\makebox[\textwidth]{\includegraphics[width=15cm]{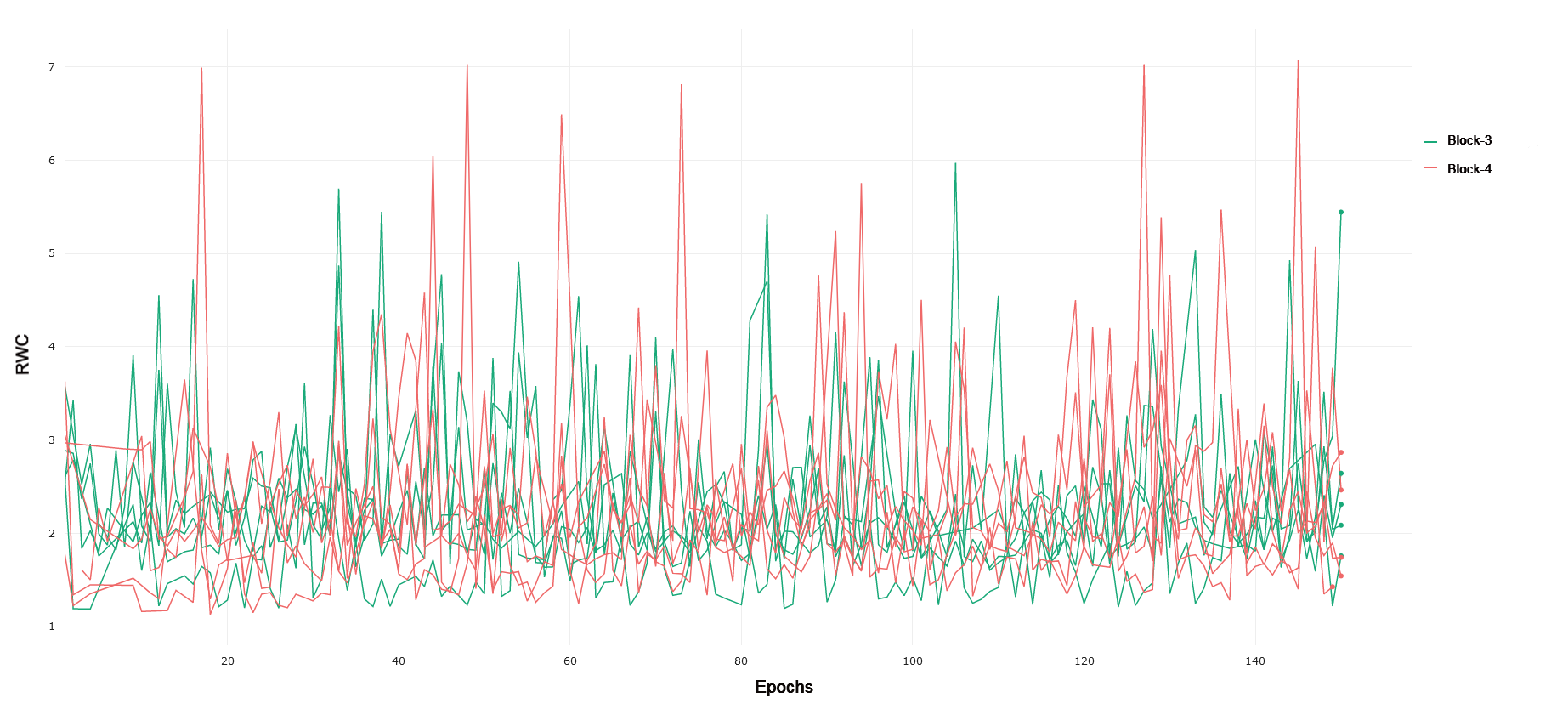}}
    \caption{RWC for ResNet-18 Blocks 3-4 on CIFAR-10}
    \label{fig:2}
\end{figure}

\textbf{CIFAR-10 and CIFAR-100} Trained on CIFAR-10, ResNet-18 exhibits an increased relative weight change in later layers of the network as compared to earlier layers. Block 1 of the network, consisting of the first 4 convolutional layers following the input convolution layer, exhibits lower relative weight change over the duration of training as compared to Block 2. This can be seen in \ref{fig:0}. Block 2 demonstrates a lower RWC as compared to Block 3, and Block 3's relative weight change exhibits similar behavior to the last block of the network. These trends can be seen in \ref{fig:1} and \ref{fig:2}, respectively. This instance of ResNet-18 achieved an accuracy of 91\% on the test set provided by the PyTorch distribution of CIFAR-10. The similar scale of relative weight change in Blocks 3 and 4, coupled with the relatively good performance of the converged network may indicate that ResNet-18 is able to learn the CIFAR-10 task without having to fully utilize the representational capacity of the last layers in the network present in Block 4. This interplay between complexity and the behavior of RWC in later layers of deep networks becomes evident in other results that follow. In general, we see that later layers demonstrate an increased RWC as compared to earlier layers in the network. 

\begin{figure}[hbt!]
    \centering
    \noindent\makebox[\textwidth]{\includegraphics[width=15cm]{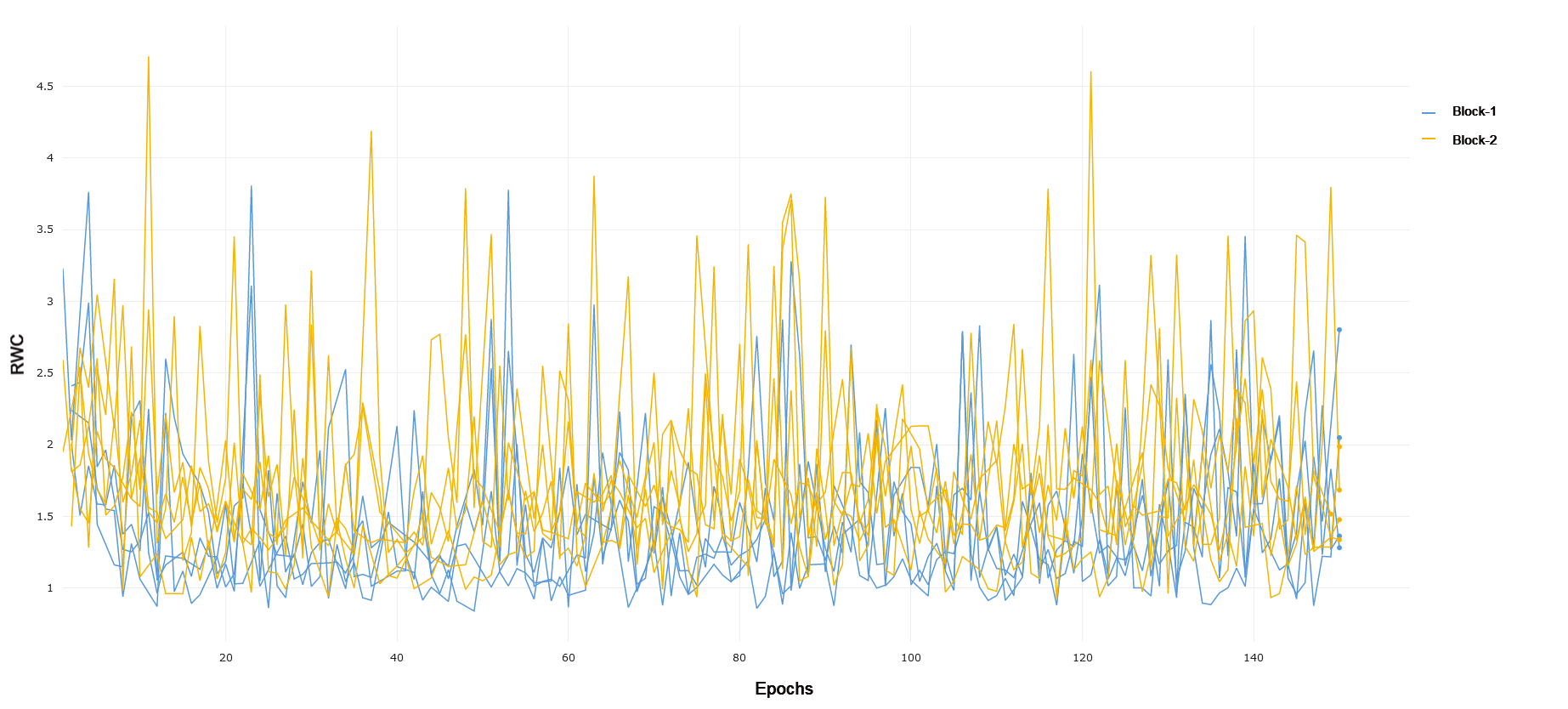}}
    \caption{RWC for ResNet-18 Blocks 1-2 on CIFAR-100}
    \label{fig:3}
\end{figure}

\begin{figure}[hbt!]
    \centering
    \noindent\makebox[\textwidth]{\includegraphics[width=15cm]{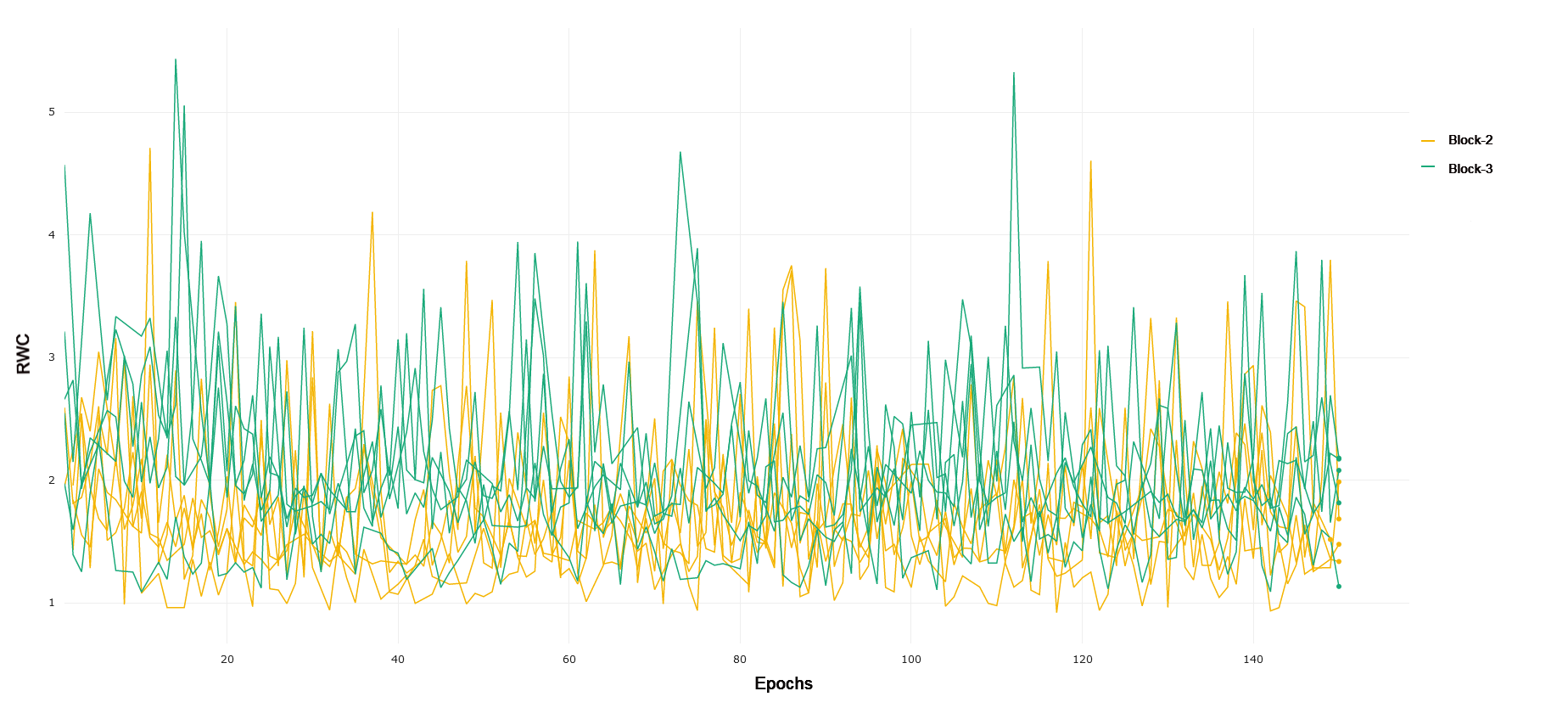}}
    \caption{RWC for ResNet-18 Blocks 2-3 on CIFAR-100}
    \label{fig:4}
\end{figure}

CIFAR-100 is a significantly more difficult task as compared to CIFAR-10, consisting of 100 classes for roughly the same number of data samples. Again, we see a trend of RWC increasing in later layers as compared to earlier layers through the course of training. Block 1 exhibits lower RWC as compared to Block 2, while Block 2 exhibits less RWC as compared to Block 3 in general. These trends can be observed in \ref{fig:3} and in \ref{fig:4}. Interestingly, there is a noticeable difference between the RWC of Block 3 and 4, with the latter having a generally higher RWC. This is in contrast to what was observed in CIFAR-10, where these blocks had similar RWC over the course of training. This difference may be the result of ResNet-18 using more of its representational capacity in later layers to solve the target task, as the CIFAR-100 task is significantly more difficult than CIFAR-10. The network achieved a 64\% classification accuracy on the PyTorch distribution of CIFAR-100, further emphasizing the challenging nature of this particular classification problem and the increased relative weight change in later layers. 


\begin{figure}[hbt!]
    \centering
    \noindent\makebox[\textwidth]{\includegraphics[width=15cm]{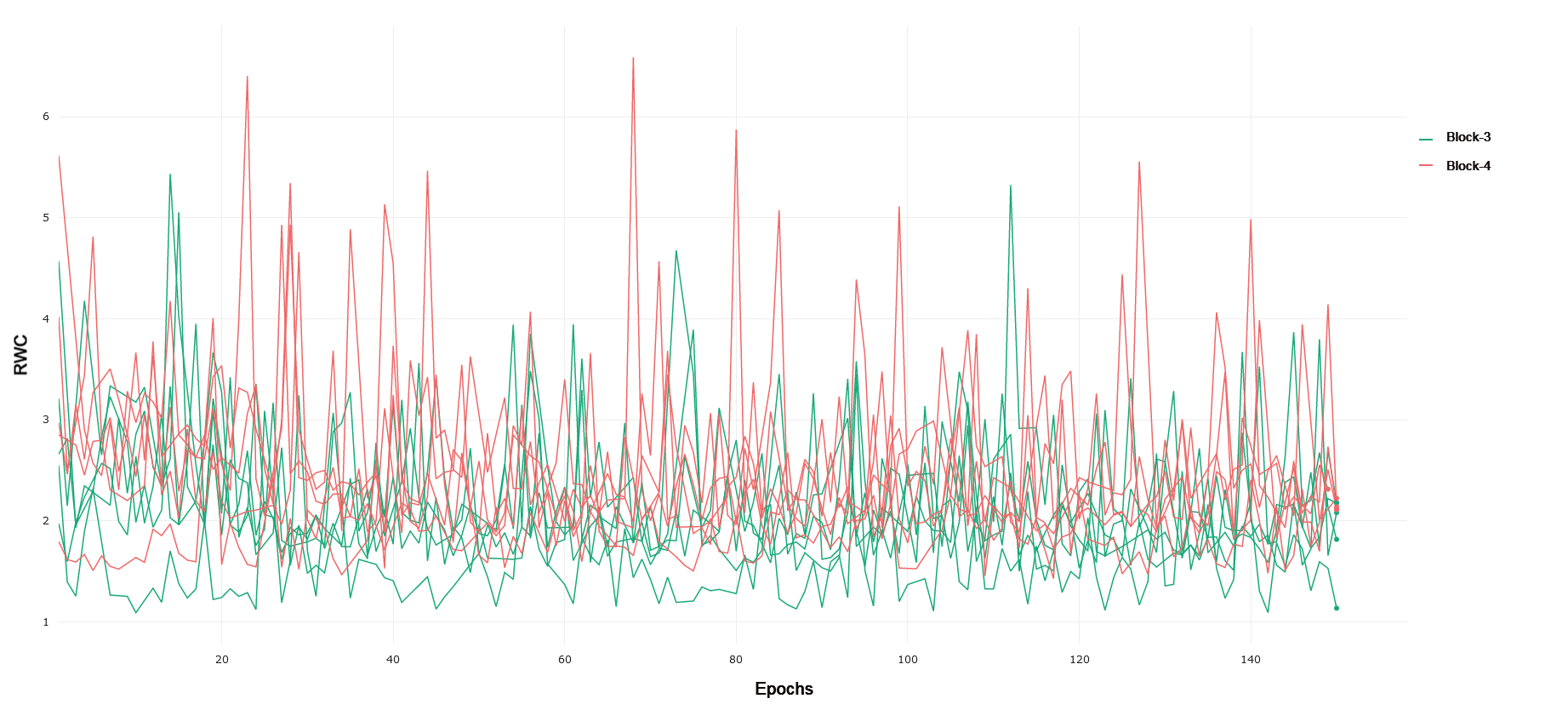}}
    \caption{RWC for ResNet-18 Blocks 3-4 on CIFAR-100}
    \label{fig:5}
\end{figure}

 \textbf{MNIST and FMNIST} Trained on MNIST, ResNet-18 exhibits a similar trend to CIFAR-10 and CIFAR-100, with RWC increasing in later Blocks as compared to earlier ones. Interestingly, we again see that Block 4 exhibits lower relative weight change as compared to Block 3, mirroring the trend seen in CIFAR-10. This, along with the fact that ResNet-18 achieves a 99\% test accuracy on MNIST, corroborates the interplay of complexity of the learning task and the capacity of the network, as MNIST is a simpler prediction problem and ResNet-18 likely does not need the full representational capacity of the layers in Block 4. FMNIST is a noticeably more difficult task with more complex images consisting of fashion objects rather than handwritten digits, and ResNet-18 achieves 92\% performance on the test set. Blocks 1 and 2 exhibit lower RWC as compared to Block 3. Block 4 again is lower than block three, reflecting the same trend as seen in CIFAR-10 and MNIST, pointing to lower recruitment of the last few layers for the task. In general, the magnitude of RWC across all layers is observably lower for MNIST as compared to FMNIST, highlighting the increased difficulty and weight adjustments required to learn a solution for the latter task.

\subsection{VGG}
We use VGG-19 with batch norm due to its analogous depth when compared to ResNet-18. VGG-19 constitutes a more traditional convolutional neural network, stacking layers by down-sampling the images passed through the network. We compared the layer-wise learning by referring to the first 4 convolutional layers as earlier layers. Layers 5 through 11 were treated as middle layers, and layer 11 onwards were treated as later layers. We chose to divide the layers in this manner after noticing a common trend in the RWC in these layers as explained in the rest of this section. 

\begin{figure}[hbt!]
    \centering
    \noindent\makebox[\textwidth]{\includegraphics[width=15cm]{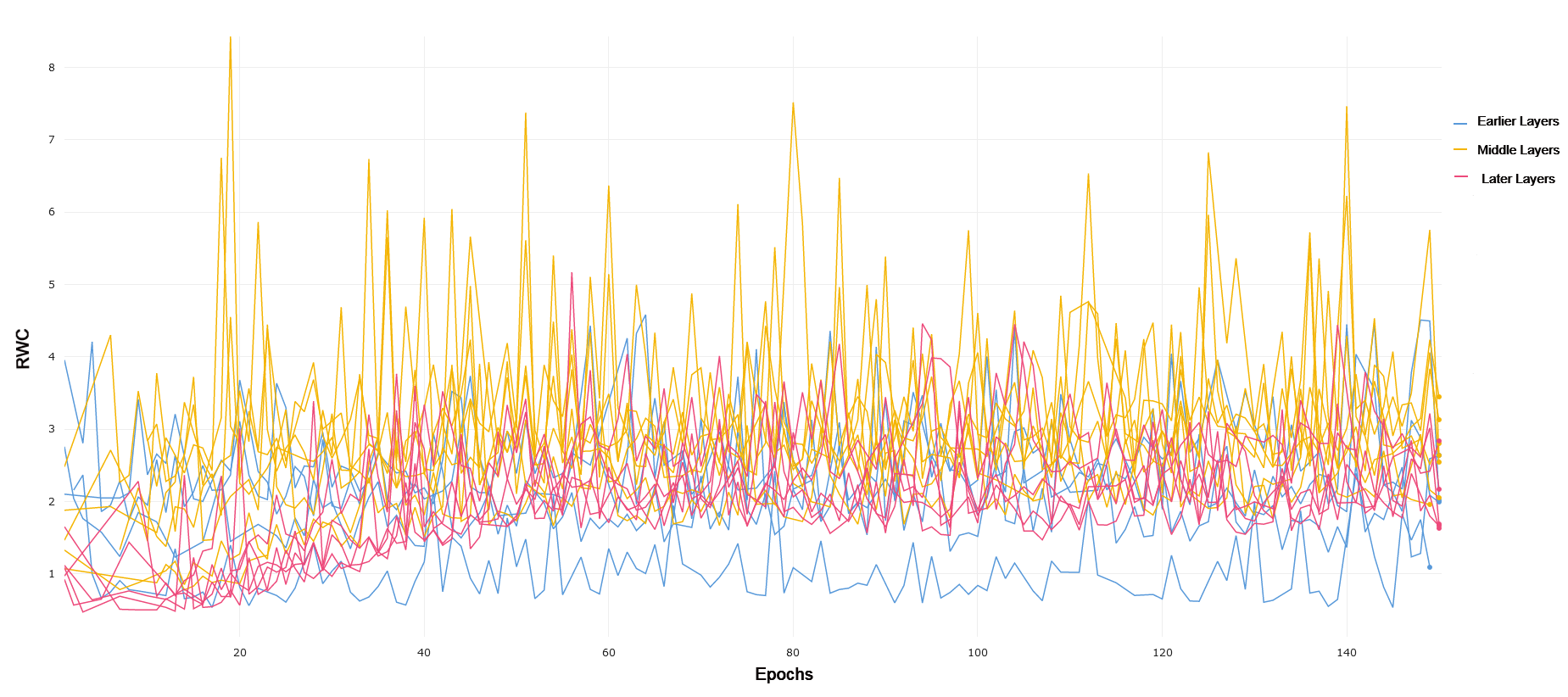}}
    \caption{RWC for VGG on CIFAR-10}
    \label{fig:6}
\end{figure}

\begin{figure}[hbt!]
    \centering
    \noindent\makebox[\textwidth]{\includegraphics[width=15cm]{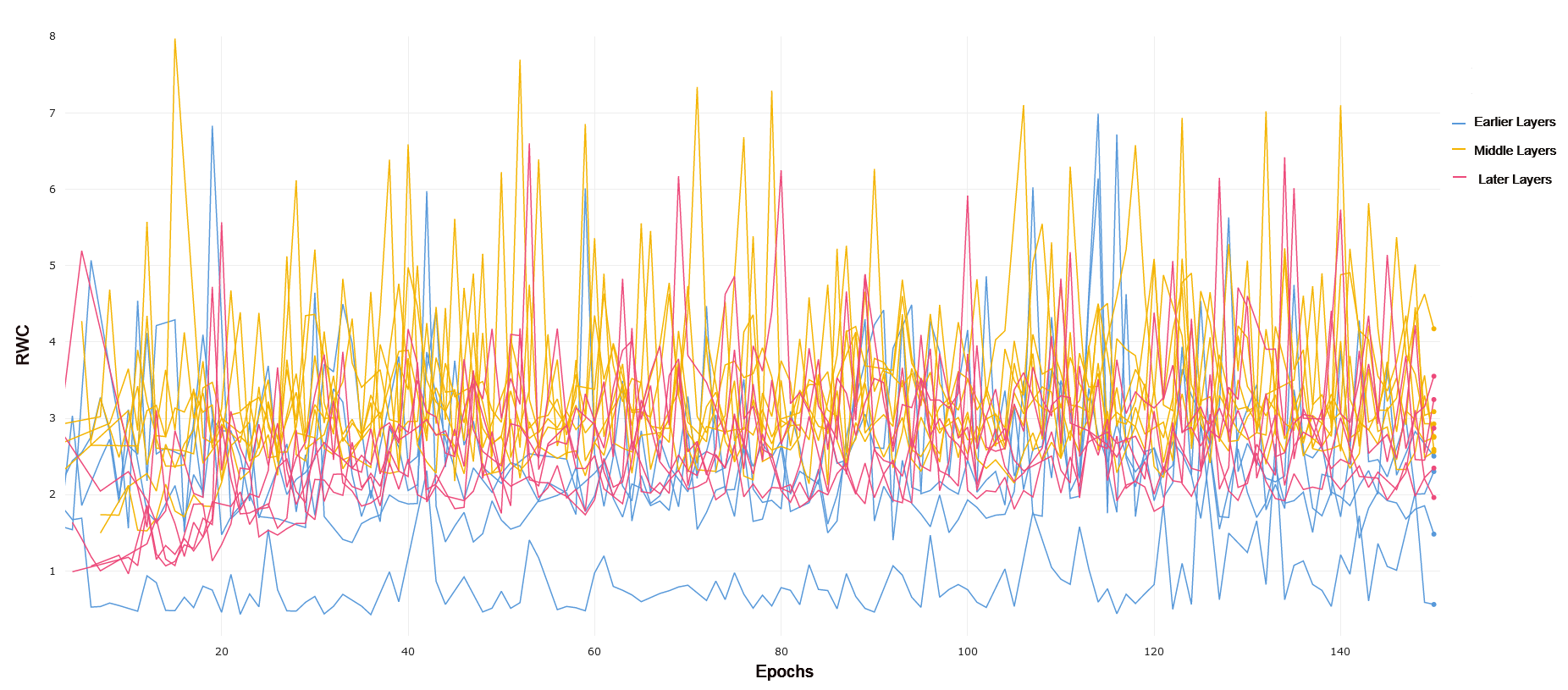}}
    \caption{RWC for VGG on CIFAR-100}
    \label{fig:7}
\end{figure}

\textbf{CIFAR-10 and CIFAR-100} VGG-19 trained on CIFAR-10 exhibits similar behavior to ResNet-18 trained on CIFAR-10, where early layers exhibit lower relative weight change than middle and later layers. This can be observed in \ref{fig:6}. Later layers again show lower relative weight change compared to the middle layers, demonstrating the same interplay of complexity and model capacity. For CIFAR-100, demonstrated in \ref{fig:7}, we see that the general RWC is higher in magnitude across all layers. This trend may be due to the difficulty in the learning task. We again see middle and later layers with higher RWC compared to early layers, though the difference between middle and later layers themselves is less pronounced, pointing to more recruitment of later layers in a similar manner to what was observed in ResNet-18 on CIFAR-100. VGG-19 achieved a 90.5\% and 63.6\% test accuracy on CIFAR-10 and CIFAR-100, respectively.


\textbf{MNIST and FMNIST} VGG-19 exhibits very similar trends in both MNIST and FMNIST, with early layers having lower RWC as compared to middle layers, but with later layers having the lowest overall RWC. Both of these learning tasks are relatively simple, and VGG-19 achieves 98.5\% and 91.6\% on MNIST and FMNIST, respectively. Lower overall RWC in later layers generally points to the same trend of deep architectures not needing to adjust learning in later layers as frequently for simpler tasks. Overall, the results and performance across both VGG-19 and ResNet-18 have a high degree of similarity. In general, the magnitude of RWC across all layers is again lower for MNIST as compared to FMNIST, similar to ResNet-18.

\subsection{AlexNet}
AlexNet is a simpler architecture as compared to VGG-19 and ResNet-18, and primarily serves as a benchmark to compare the layer-wise learning dynamics for a shallower convolutional network. The first convolutional layer is referred to as the early layer. The second and the third layers are the middle layers and the remaining 2 layers are referred to as later layers. Figures for AlexNet are included in the supplementary materials, and general observed trends are covered here. 

\textbf{CIFAR-10 and CIFAR-100} Trained on CIFAR-10, AlexNet exhibits the same trends as the other networks, with early layers and later layers exhibiting lower overall RWC as compared to middle layers. This trend can be seen in figure 9 in the supplementary material. AlexNet demonstrates increasing RWC when trained on CIFAR-100, with early layer exhibiting less RWC as compared to middle layers and middle layers exhibiting less RWC as compared to later layers. This is consistent with the behavior observed in ResNet-18 when trained on CIFAR-100, and may point to increased utilization of capacity in the network for a much harder task. AlexNet achieved an 81\% test accuracy on CIFAR-10, and a 53\% test accuracy on CIFAR-100. These performances are to be expected, as AlexNet is a much shallower network. 

\textbf{MNIST and FMNIST} AlexNet trained on MNIST and FMNIST generally shows a trend of earlier layers having a lower RWC as compared to middle layers, and middle layers having a lower RWC as compared to later layers. In the same framework focused on the interplay between model capacity and task complexity, it seems that AlexNet uses its later layers' representational capacity for both MNIST and FMNIST. AlexNet achieves 98.5\% on MNIST and 89.5\% on FMNIST, respectively.

\section{Conclusion}
In general, we see that relative weight change increases in later layers as compared to earlier ones across the different convolutional architectures, both deep and shallow, and across the different classification tasks. An interesting general trend emerges when networks are trained on comparatively simpler tasks like CIFAR-10 and MNIST, where later layers exhibit lower RWC as compared to middle layers of the network. On more complex tasks like CIFAR-100, we see that later layers exhibit higher RWC compared to early and middle layers, potentially indicating  increased usage of the representational capacity of the network. Understanding layer-wise learning dynamics in deep networks provides a promising and impactful avenue of research, and has several potential future directions. These include the design of alternative metrics for layer-wise and neuron-wise learning in deep networks, pruning and freezing methods based on these metrics, and the empirical assessment of these metrics on other problem domains, including Natural Language Processing, Speech Recognition, and Reinforcement Learning.

\medskip

\bibliography{refs}
\bibliographystyle{unsrt}
\section{Appendix}
\begin{table}[hbt!]
\centering
\caption{Detailed hyperparameters used for training}
\label{tab:training-hyperparameters}
\begin{tabular}{|c|c|c|c|c|}
\hline
\textbf{Architecture}      & \textbf{Datasets} & \textbf{LR} & \textbf{Momentum} & \textbf{Weight Decay} \\ \hline
\multirow{4}{*}{ResNet18}  & CIFAR-10          & 0.1         & 0.9               & 0.0001                \\ \cline{2-5} 
                           & CIFAR-100         & 0.1         & 0.9               & 0.0001                \\ \cline{2-5} 
                           & MNIST             & 0.1         & 0.9               & 0.0001                \\ \cline{2-5} 
                           & FMNIST            & 0.1         & 0.9               & 0.0001                \\ \hline
\multirow{4}{*}{VGG19\_bn} & CIFAR-10          & 0.05        & 0.9               & 0.0005                \\ \cline{2-5} 
                           & CIFAR-100         & 0.05        & 0.9               & 0.0005                \\ \cline{2-5} 
                           & MNIST             & 0.05        & 0.9               & 0.0005                \\ \cline{2-5} 
                           & FMNIST            & 0.05        & 0.9               & 0.0005                \\ \hline
\multirow{4}{*}{AlexNet}   & CIFAR-10          & 0.001       & 0.9               & 0.0001                \\ \cline{2-5} 
                           & CIFAR-100         & 0.01        & 0.9               & 0.0001                \\ \cline{2-5} 
                           & MNIST             & 0.1         & 0.9               & 0.0001                \\ \cline{2-5} 
                           & FMNIST            & 0.1         & 0.9               & 0.0001                \\ \hline
\end{tabular}
\end{table}
\subsection{RWC Plots}
\begin{figure}[hbt!]
    \centering
    \noindent\makebox[\textwidth]{\includegraphics[width=15cm]{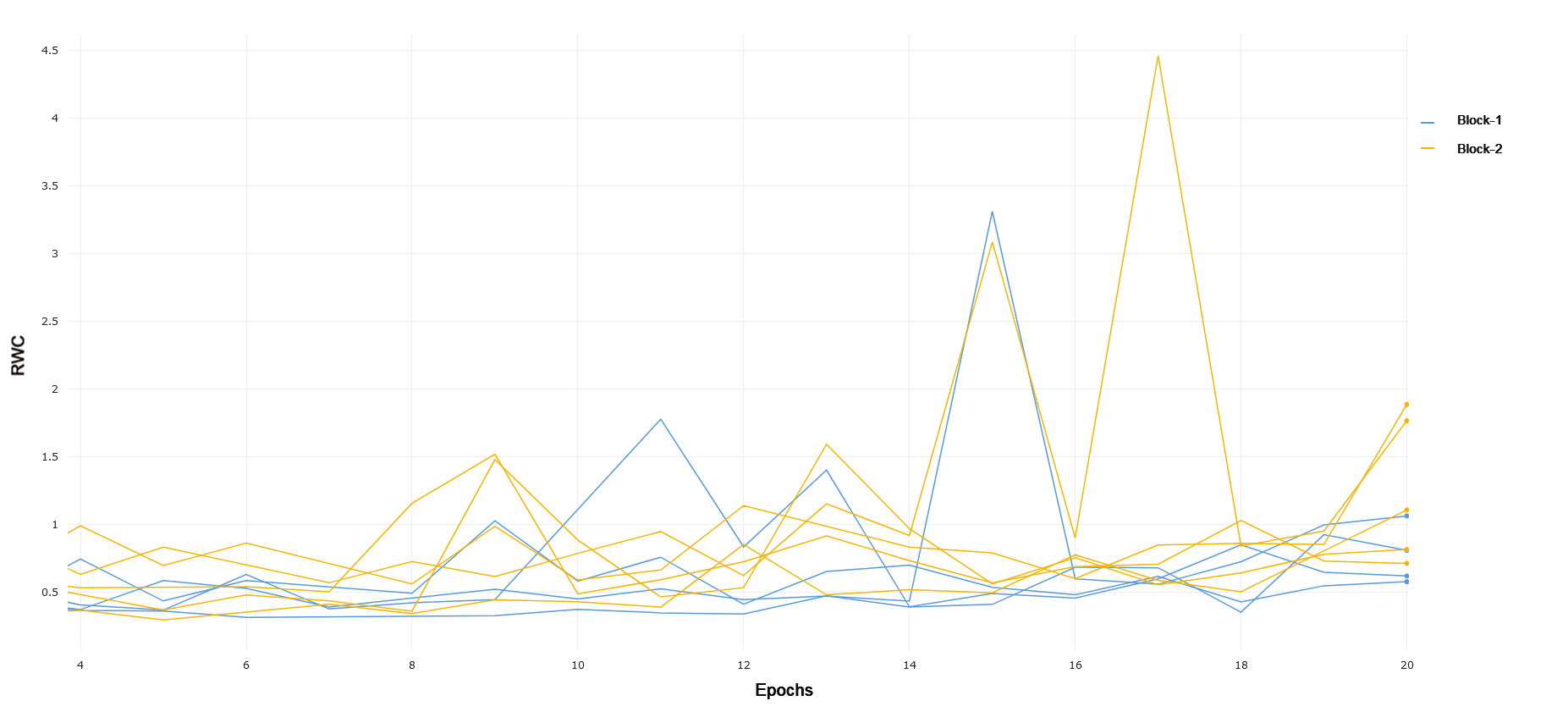}}
    \caption{RWC for Resnet18 Blocks 1-2 on MNIST}
    \label{fig:8}
\end{figure}
\begin{figure}[hbt!]
    \centering
    \noindent\makebox[\textwidth]{\includegraphics[width=15cm]{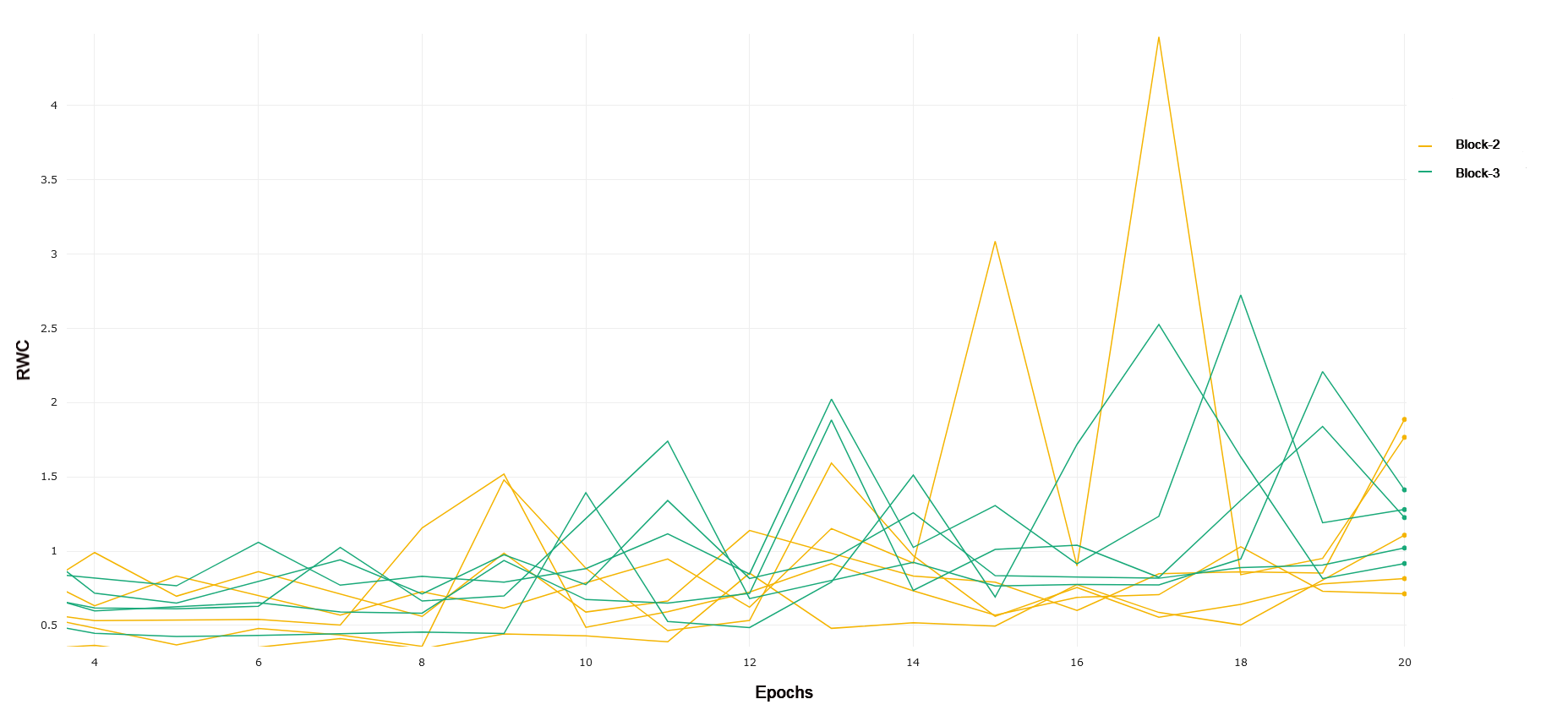}}
    \caption{RWC for Resnet18 Blocks 2-3 on MNIST}
    \label{fig:9}
\end{figure}
\begin{figure}[hbt!]
    \centering
    \noindent\makebox[\textwidth]{\includegraphics[width=15cm]{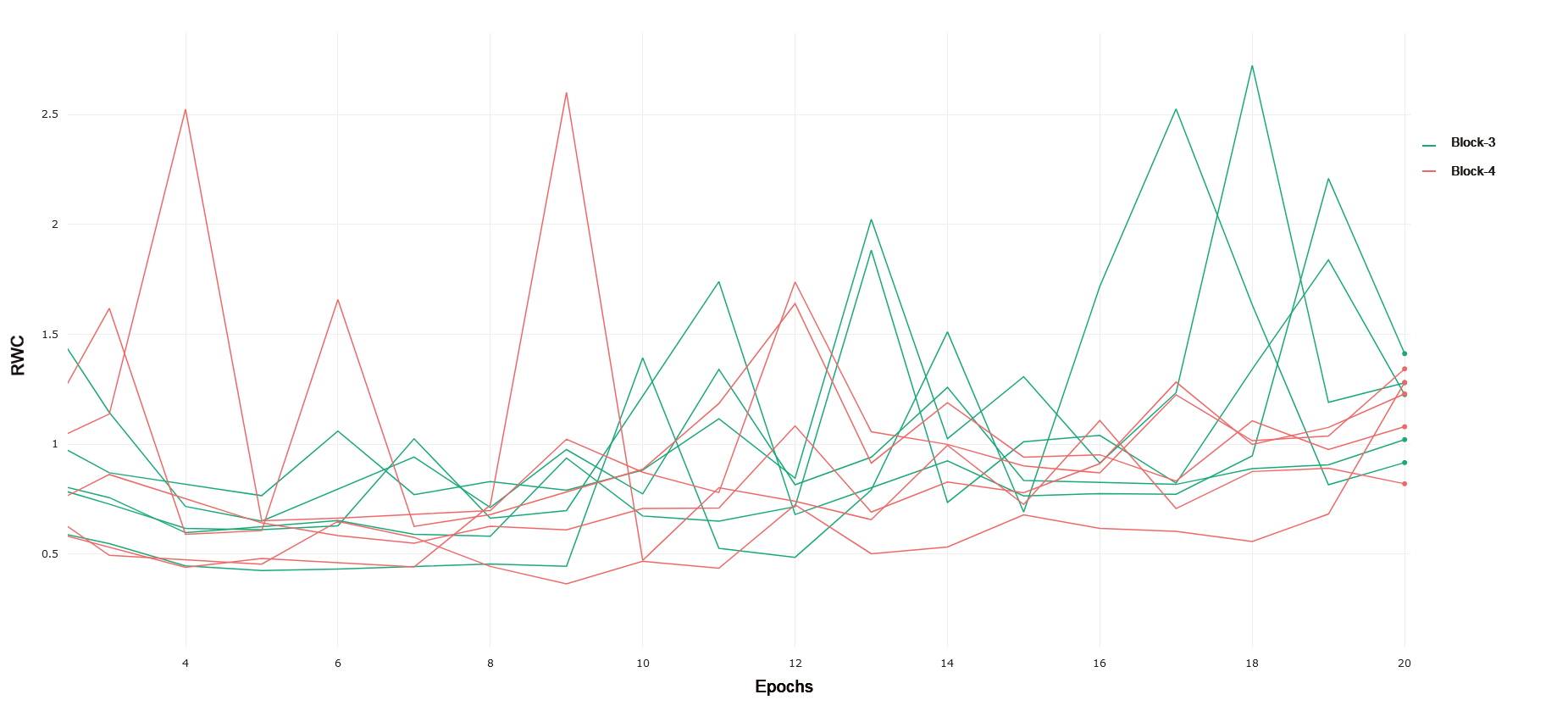}}
    \caption{RWC for Resnet18 Blocks 3-4 on MNIST}
    \label{fig:10}
\end{figure}
\begin{figure}[hbt!]
    \centering
    \noindent\makebox[\textwidth]{\includegraphics[width=15cm]{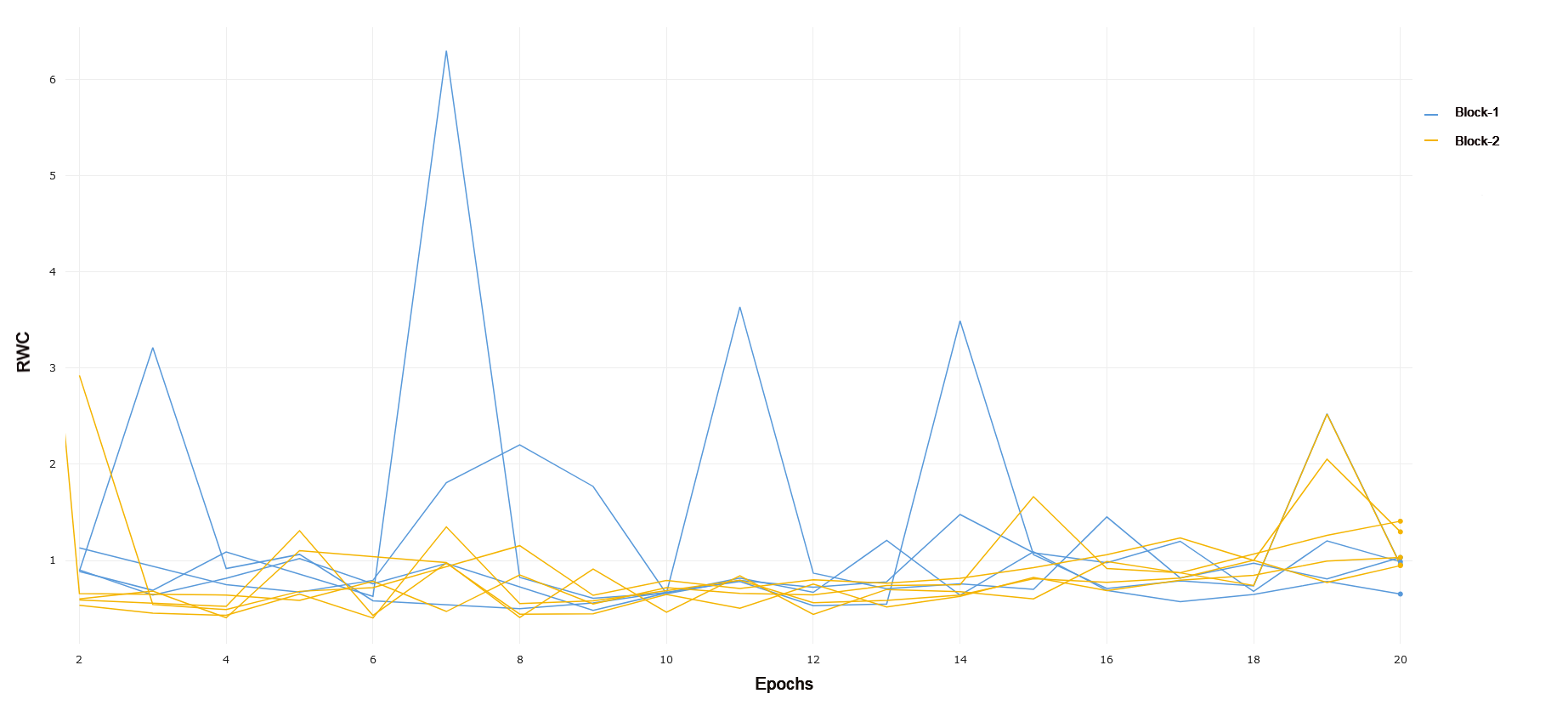}}
    \caption{RWC for ResNet-18 Blocks 1-2 on FashionMNIST}
    \label{fig:11}
\end{figure}
\begin{figure}[hbt!]
    \centering
    \noindent\makebox[\textwidth]{\includegraphics[width=15cm]{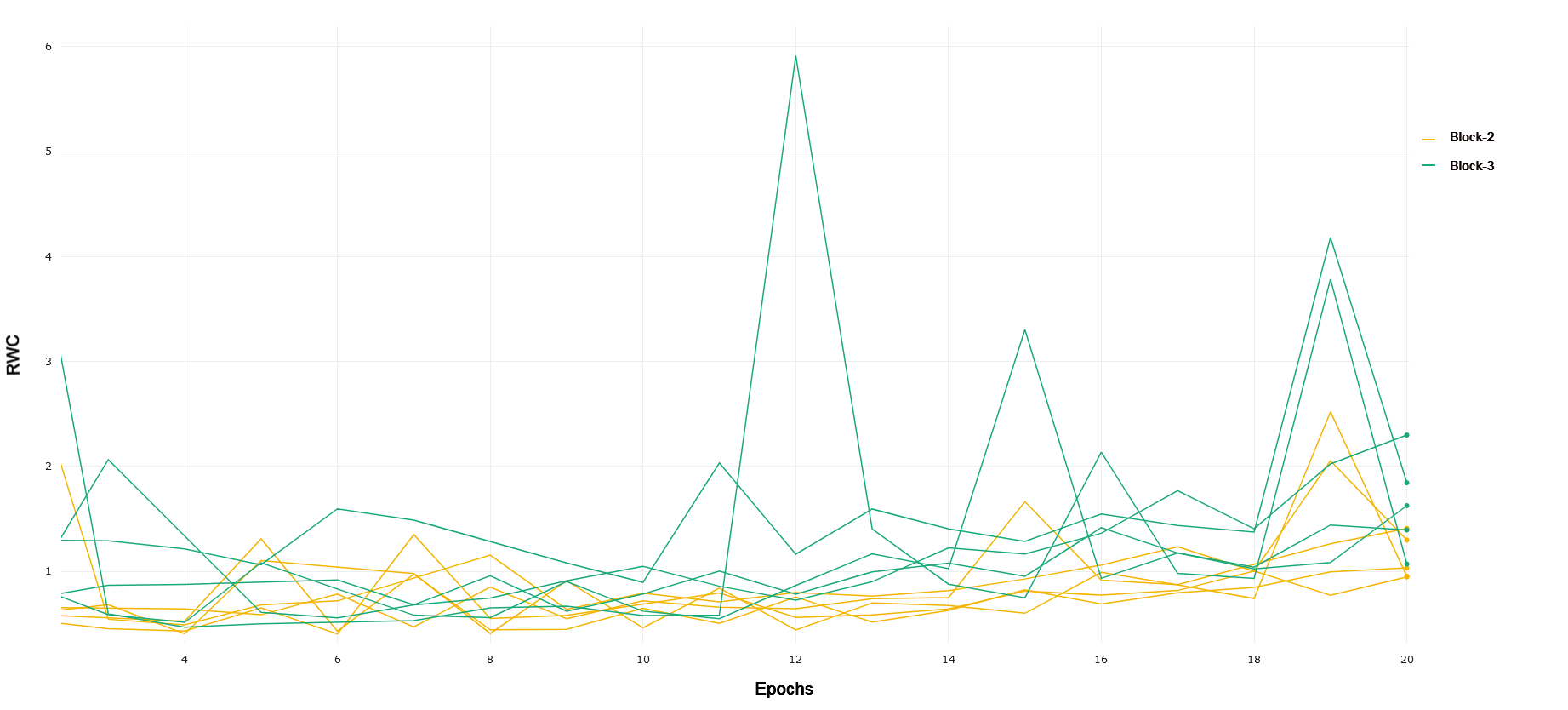}}
    \caption{RWC for ResNet-18 Blocks 2-3 on FashionMNIST}
    \label{fig:12}
\end{figure}
\begin{figure}[hbt!]
    \centering
    \noindent\makebox[\textwidth]{\includegraphics[width=15cm]{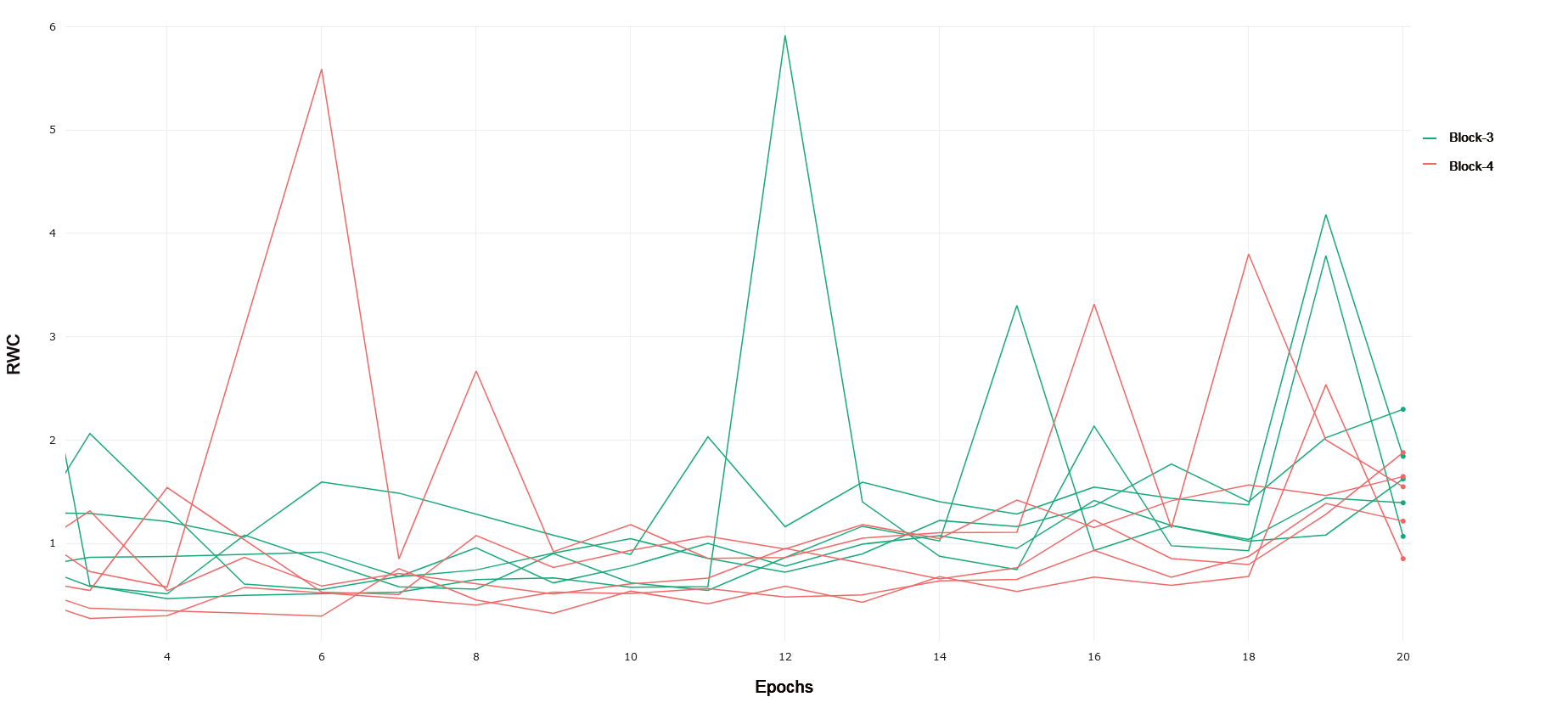}}
    \caption{RWC for ResNet-18 Blocks 3-4 on FashionMNIST}
    \label{fig:13}
\end{figure}
\begin{figure}[hbt!]
    \centering
    \noindent\makebox[\textwidth]{\includegraphics[width=15cm]{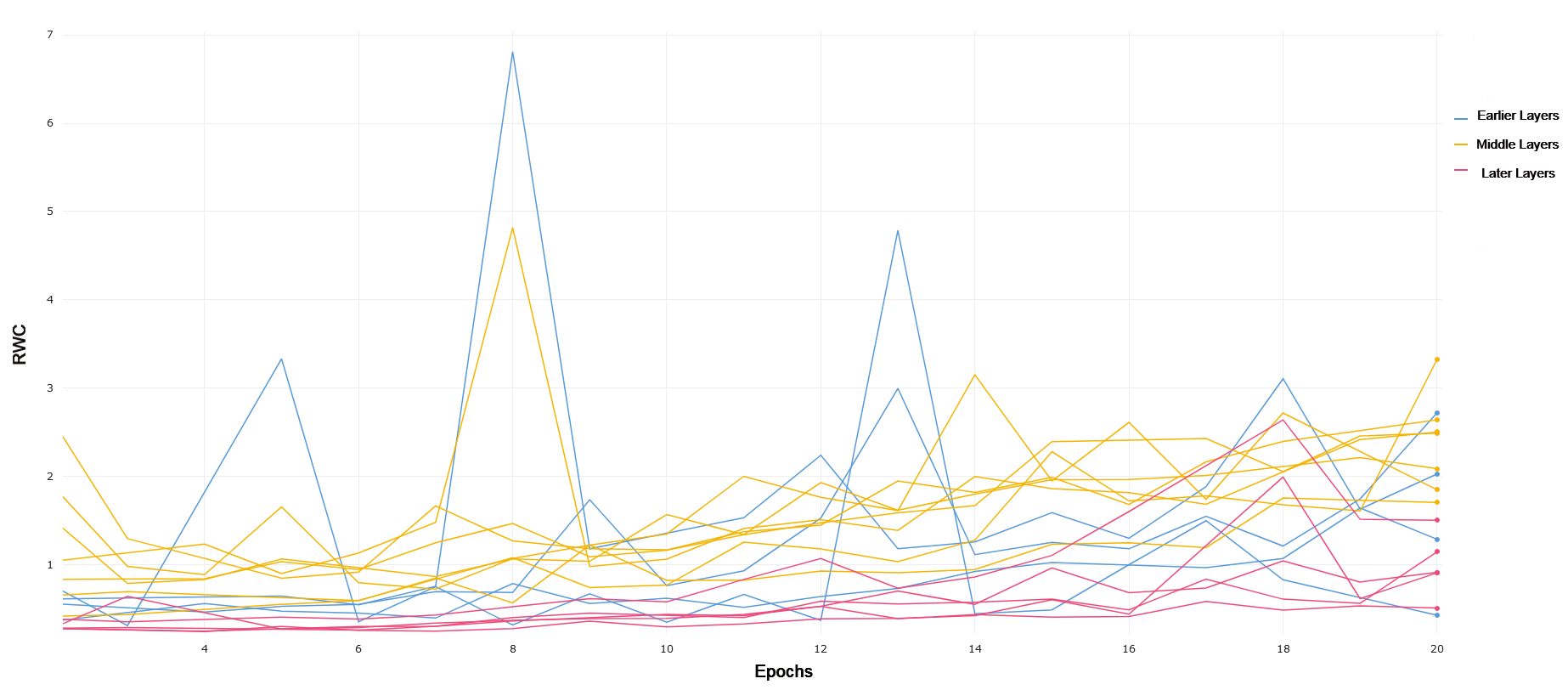}}
    \caption{RWC for VGG on MNIST}
    \label{fig:14}
\end{figure}
\begin{figure}[hbt!]
    \centering
    \noindent\makebox[\textwidth]{\includegraphics[width=15cm]{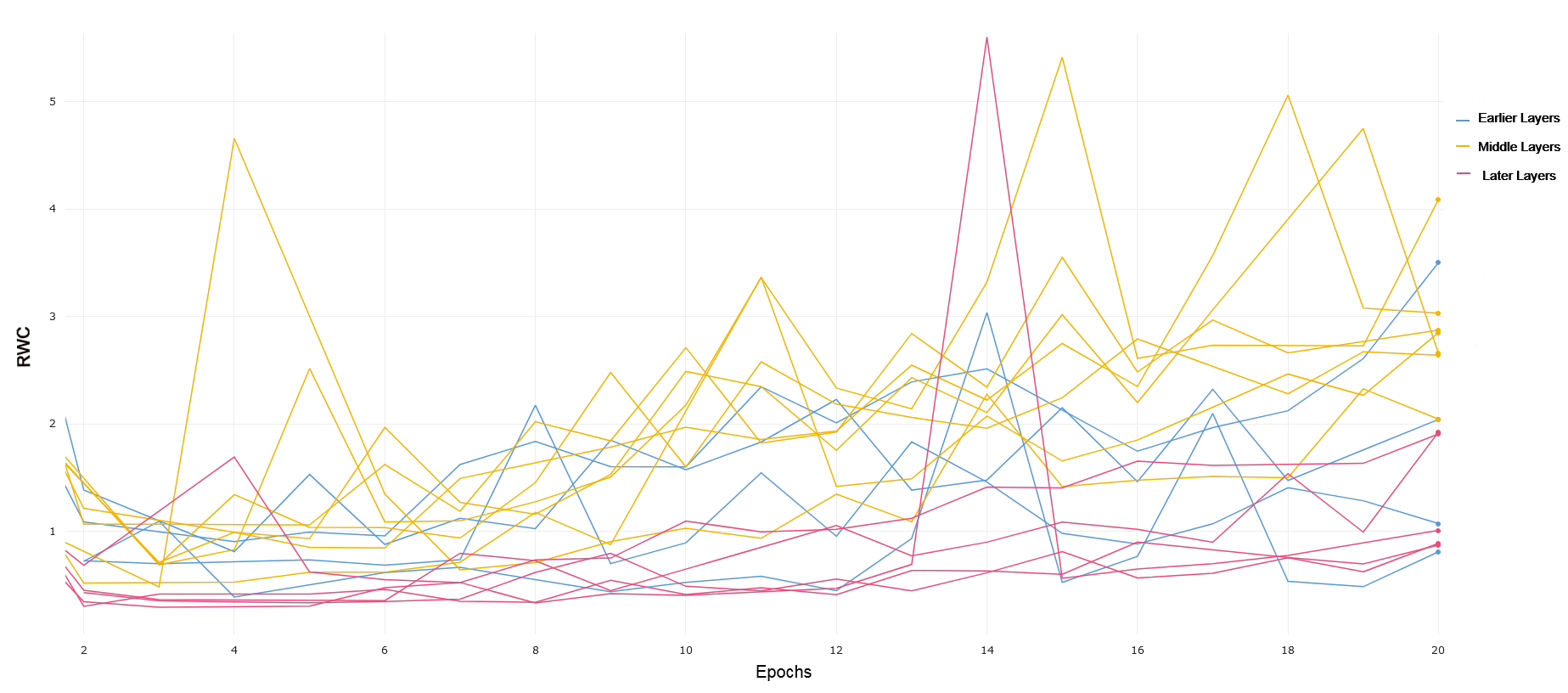}}
    \caption{RWC for VGG on FashionMNIST}
    \label{fig:15}
\end{figure}
\begin{figure}[hbt!]
    \centering
    \noindent\makebox[\textwidth]{\includegraphics[width=15cm]{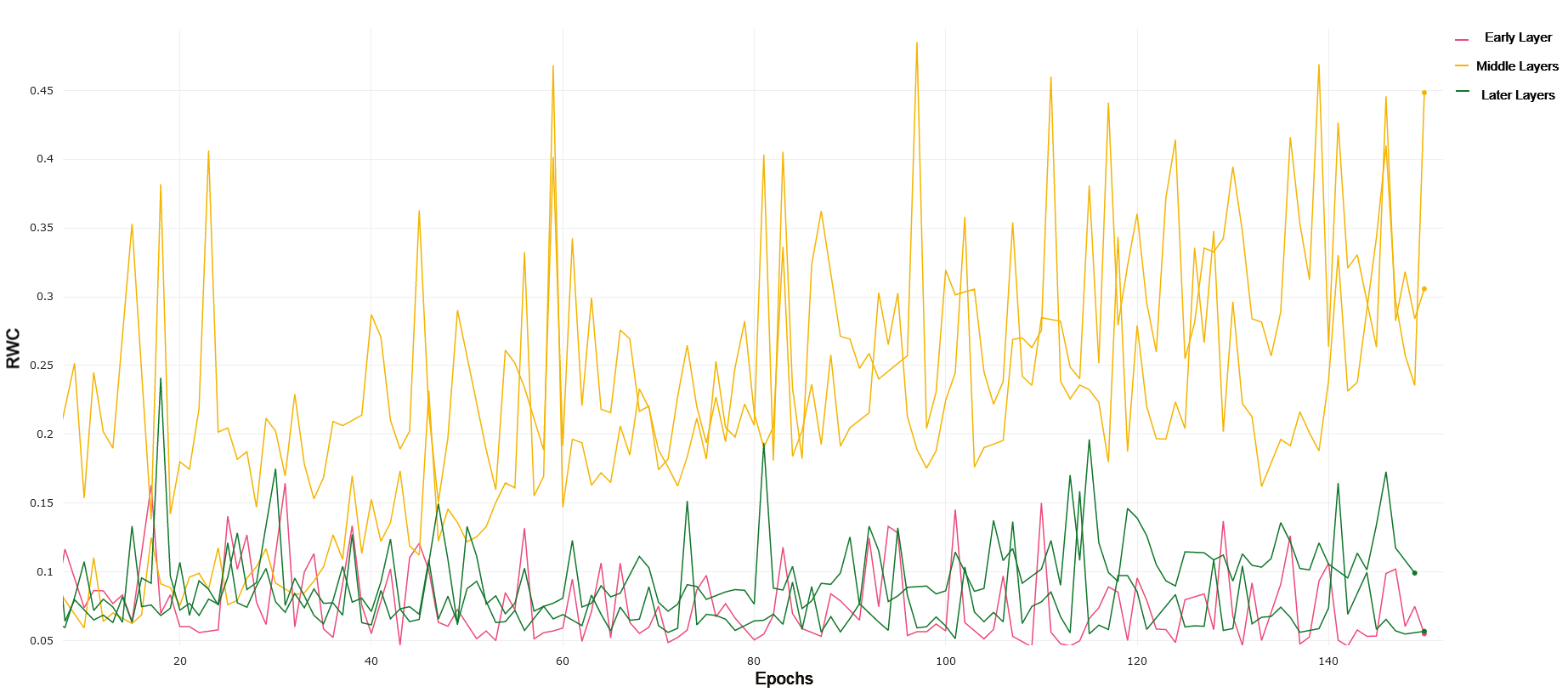}}
    \caption{RWC for Alexnet on CIFAR10}
    \label{fig:16}
\end{figure}
\begin{figure}[hbt!]
    \centering
    \noindent\makebox[\textwidth]{\includegraphics[width=15cm]{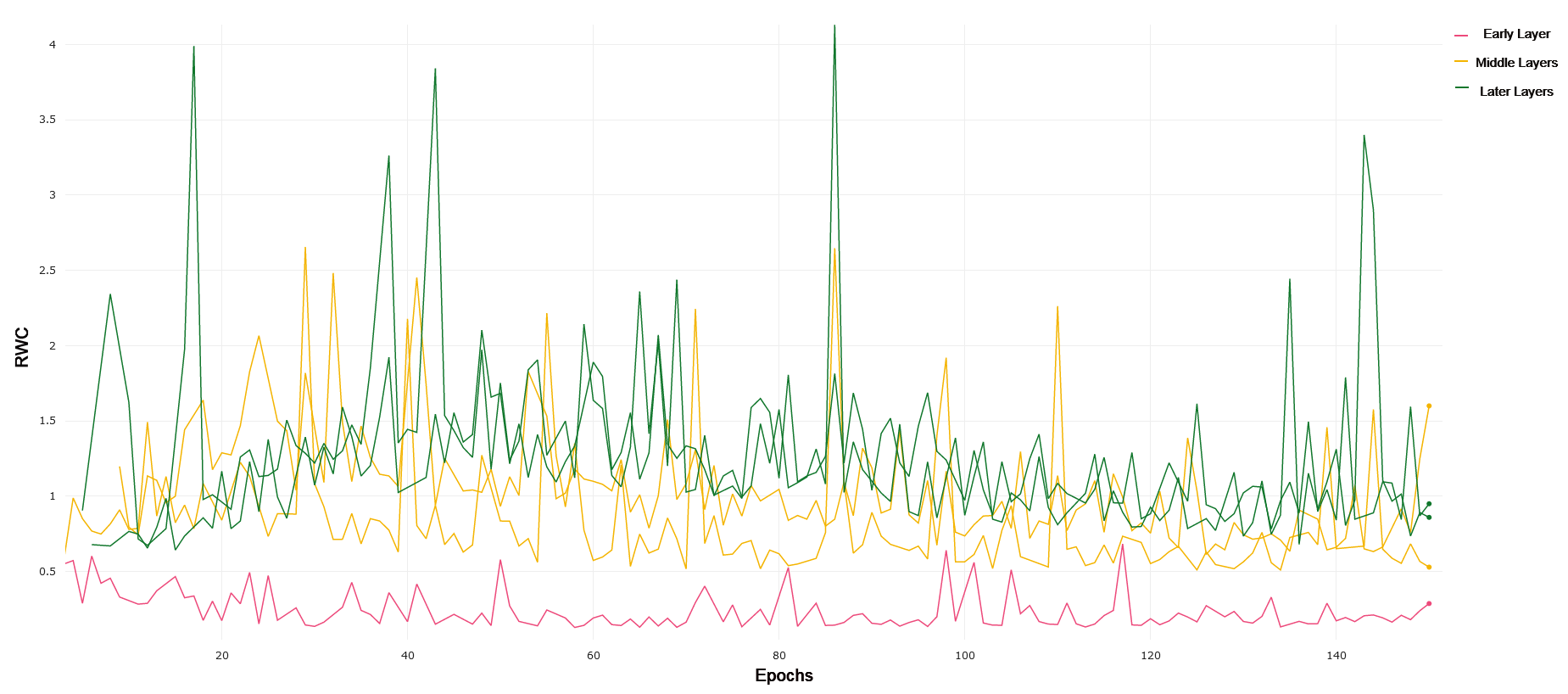}}
    \caption{RWC for Alexnet on CIFAR100}
    \label{fig:17}
\end{figure}
\begin{figure}[hbt!]
    \centering
    \noindent\makebox[\textwidth]{\includegraphics[width=15cm]{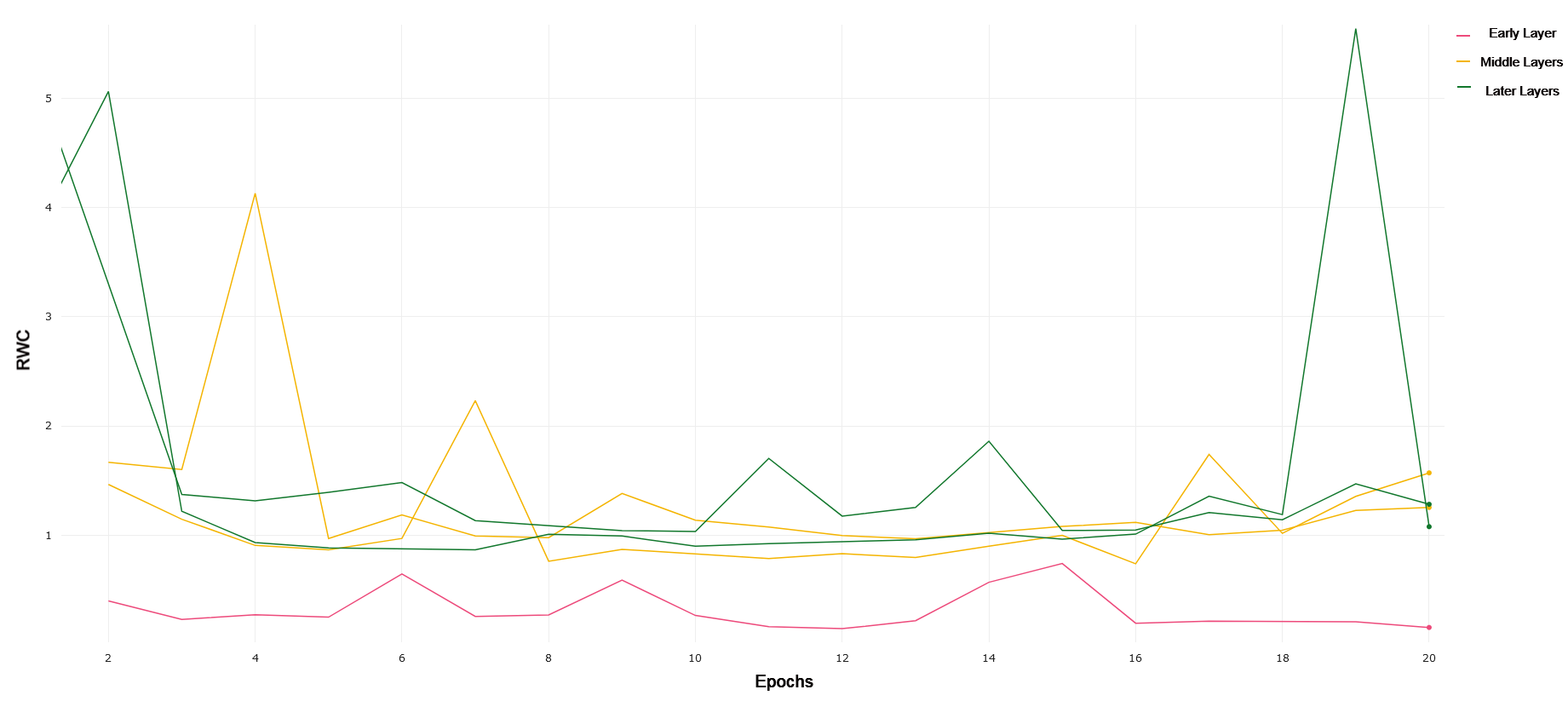}}
    \caption{RWC for Alexnet on MNIST}
    \label{fig:18}
\end{figure}
\begin{figure}[hbt!]
    \centering
    \noindent\makebox[\textwidth]{\includegraphics[width=15cm]{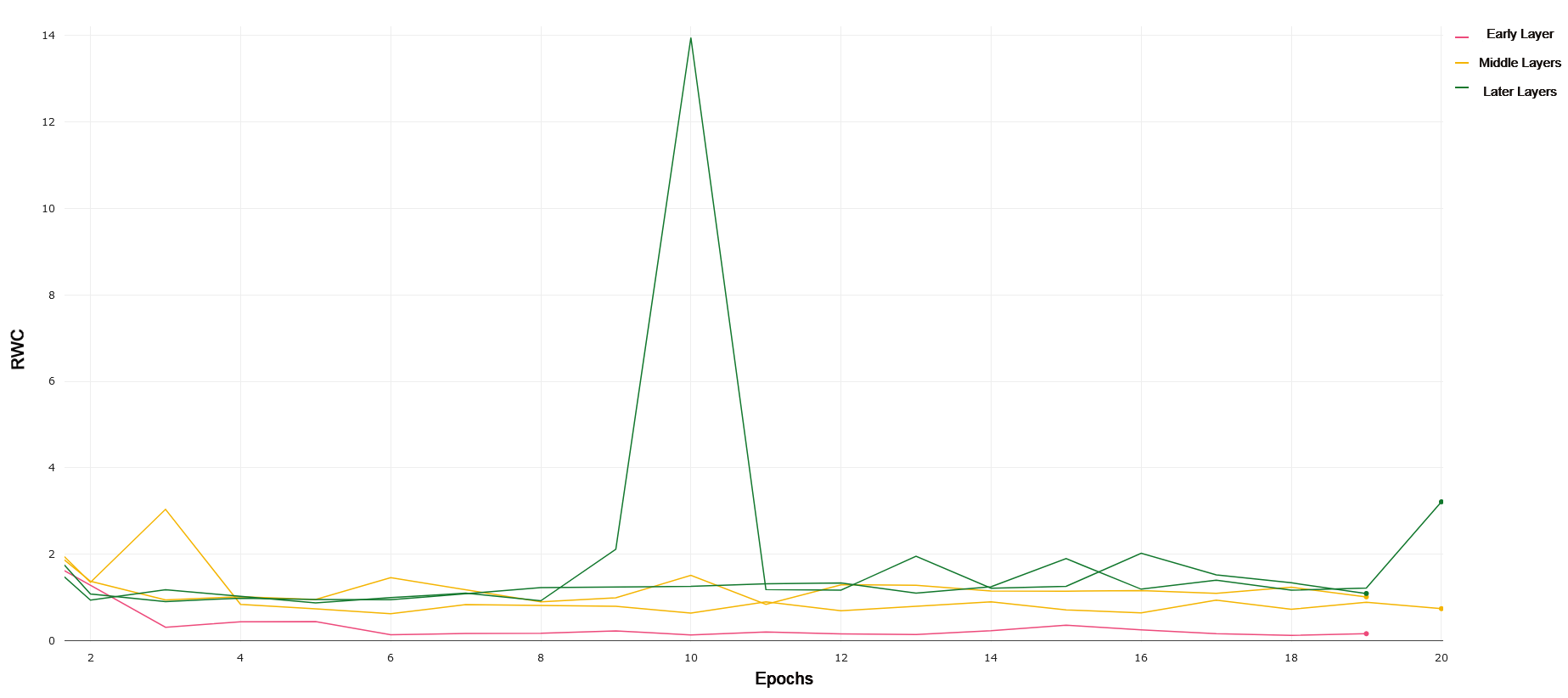}}
    \caption{RWC for Alexnet on FashionMNIST}
    \label{fig:19}
\end{figure}

\end{document}